%% file: main.tex
\documentclass{article} %
\usepackage{iclr2025_conference,times}

\usepackage{hyperref}
\usepackage{url}
\usepackage{enumitem}
\usepackage{bbm}

\title{
Improving the evaluation of samplers on multi-modal targets
}

\author{Louis Grenioux\textsuperscript{$\ast$} \& Maxence Noble\thanks{Both authors contributed equally.} \\
CMAP, CNRS \\
École polytechnique, Institut Polytechnique de Paris\\
91120 Palaiseau, France \\
\texttt{\{louis.grenioux,maxence.noble-bourillot\}@polytechnique.edu} \\
\And
Marylou Gabrié \\
Laboratoire de Physique de l’École normale supérieure \\
ENS, Université PSL, CNRS, Sorbonne Université, Université de Paris \\
75005 Paris, France \\
\texttt{marylou.gabrie@ens.fr}
}

\iclrfinalcopy %

\usepackage{mathrsfs}  
\usepackage{amsmath,amsfonts,bm}
\usepackage{cleveref}
\usepackage{graphicx}
\usepackage{comment}
\usepackage{soul}

\def\rset{\mathbb{R}}
\def\normcte{\mathcal{Z}}
\def\densityGaussian{\mathrm{N}}
\def\eqsp{\;}
\def\rmd{\mathrm{d}}
\newcommand{\1}{\mathbbm{1}}
\def\Idd{\mathrm{I}_d}

\newcommand{\abs}[1]{\left\lvert#1\right\rvert}

\begin{document}

\maketitle

\begin{abstract}
Addressing multi-modality constitutes one of the major challenges of sampling. In this reflection paper, we advocate for a more systematic evaluation of samplers towards two sources of difficulty that are mode separation and dimension. For this, we propose a synthetic experimental setting that we illustrate on a selection of samplers, focusing on the challenging criterion of recovery of the mode relative importance. These evaluations are crucial to diagnose the potential of samplers to handle multi-modality and therefore to drive progress in the field.
\end{abstract}

\input{1_intro}

\input{2_metrics}

\input{3_samplers}

\input{4_conclusion}

\newpage
\section*{Acknowledgments}
We would like to thank Alain Durmus for many stimulating discussions and for reviewing the manuscript prior to submission. LG acknowledge funding from Hi! Paris. This work was partially performed using HPC resources from GENCI–IDRIS (AD011015234).

\bibliography{main}
\bibliographystyle{iclr2025_conference}

\newpage
\appendix
\input{0_appendix}

\end{document}

%% file: 1_intro.tex
\section{Introduction}

Consider the problem of sampling from a
probability density known up to a normalizing constant. More precisely, let $\pi$ be a target distribution  with probability density given by 
$x \mapsto  \gamma(x)   / \normcte$,
where $\gamma: \rset^d \to \rset_+$ can be pointwise evaluated and the normalizing constant $\normcte=\int_{\rset^d}\gamma(x)\rmd x$ is intractable.
This reflection paper is concerned with 
cases where the distribution $\pi$ is multi-modal, i.e., $\gamma$ admits several local maxima. 
Such high-dimensional multi-modal distributions appear 
in various fields. In physics, they describe the thermodynamics of metastable systems, or the action of quantum field theories \citep{wolffCriticalSlowing1990}. Bayesian modeling can also face multi-modal posterior distributions in cases where the observations fail to lift ambiguities between several distinctive sets of parameters \citep{yaoStackingNonmixingBayesian2022}. Hence, continuing efforts have been made to design sampling strategies that handle multi-modality. However, sampling from multi-modal distributions is notoriously difficult, especially in large dimension $d$. %

There are two major sources of difficulties to overcome: (i) locating the modes, that is the different high-probability regions of $\pi$ induced by its local maxima, and (ii) estimating their relative importance.
Performing an exhaustive search of modes is equivalent to the problem of finding the local maxima of the possibly high-dimensional function $\gamma$, which is NP-hard \cite{pardalosCheckingLocalOptimality1988}. Although a flurry of heuristics have been developed to explore probability landscapes and find modes \citep{fletcherPracticalMethodsOptimization2000, nocedalNumericalOptimization2006}, this fact sets the problem of sampling from multi-modal distributions as arbitrarily hard in general. 

In fact, even when the location of the modes is known, the second difficulty of estimating their relative proportion remains a significant challenge to this day.
The ability of samplers to represent the different modes with correct proportions is, however, seldom reported, with exceptions including \cite{grenioux2023onsampling, blessing2024beyond, shi2024diffusionpinnsampler, noble2024learned}.
\newpage
In this reflection paper, we wish to advocate for:
\vspace{-0.2cm}
\begin{enumerate}[wide,noitemsep,label=$\bullet$]
    \item Running systematic experiments on toy mixture distributions with increasing dimension and mode separation, as these two parameters are commonly mediating the hardness of the task for common sampling technics.
    \vspace{0.1cm}
    \item Evaluation based on the criterion of accurate estimation of the statistical weight of each mode, possibly assuming knowledge of mode locations. The reasons are twofold: first because it is an interpretable metric, incurring almost no additional computational cost, and second precisely because it is a challenging test.
\end{enumerate}
Running these experiments and evaluation would enable diagnosing strengths and weaknesses of newly proposed algorithms, which is crucial to foster the development of robust algorithms in the multi-modal setting, and therefore drive progress in the field.

In the following, we start with some reminders in \Cref{sec:mode_weight} on the difficulty of sampling from multi-modal distributions, and in particular in estimating relative importance of modes. In \Cref{sec:3_samplers}, we examine the strengths and weaknesses of common sampling strategies in multi-modal scenarios and illustrate the importance of reporting performance within the proposed experimental framework.

%% file: 2_metrics.tex
\section{What is hard in sampling from multi-modal distributions?}\label{sec:mode_weight}

As mentioned in the introduction, the first challenge in sampling from a multi-modal distribution is identifying all its modes. We do not discuss this issue further, as exhaustive searches can be arbitrarily complex. Moreover, even when mode locations are known, accurate sampling remains difficult and practically relevant. In some cases, heuristics or expert knowledge can help identify modes. For instance, in physics and drug design \citep{jorgensenManyRolesComputation2004}, determining molecular binding events involves two modes: the bonded complex and the unbound molecules.

\paragraph{Benefits and pitfalls of local samplers.} 
If the target’s mode locations are known, it may seem that the sampling problem is solved by using this prior information to initialize a Markov Chain Monte Carlo (MCMC) sampler.
MCMCs based on local updates are among the most widely used sampling approaches, including Random Walk Metropolis-Hastings (RWMH) \citep{metropolis1953equation}, Metropolis-Adjusted Langevin Algorithm (MALA) \citep{roberts1996exponential}, Hamiltonian Monte Carlo (HMC) \citep{duane1987hybrid,brooks2011handbook}, and No U-Turn Sampler (NUTS) \citep{hoffman2014nuts}. These algorithms have demonstrated both theoretical and practical efficiency when the target density satisfies log-concavity assumptions \citep{dalalyan2017theoretical,durmus2019high,dwivedi2018logconcave,bourabee2020coupling}.
Paradoxically, the very properties that make these samplers effective for log-concave distributions hinder their performance in the multi-modal case, where log-concavity does not hold. While MCMC transition kernels are designed to guide the chain toward target modes, they also, with high probability, trap it within a mode, preventing broader exploration. Consequently, initializing an MCMC sampler at the modes of $\pi$—where $\gamma$ is locally log-concave—may yield locally accurate samples while failing to capture the global mode weights of $\pi$.

\paragraph{Mode weight estimation.}

To 
define the mode weights of the distribution of interest $\pi$, one can leverage a partition of its support (taken equal to $\rset^d$ here for simplicity) that draws a one-to-one correspondence between the target modes and space subsets.
Then, the mode weights of $\pi$ can be defined as the relative statistical weights of the constituents of this partition with respect to $\pi$. 
Formally, assume that $\pi$ has $K$ modes and that we have access to a partition $\{S_1, \ldots, S_K\}$ of $\rset^d$, where $S_k \subset \rset^d$ represents a region including the $k$-th mode of $\pi$. The (un-normalized) $k$-th mode weight of $\pi$ is simply defined by
\begin{align}\label{eq:mode_weight}
    w_{k} = \int_{\rset^d}\1_{S_k}(x) \gamma(x) \rmd x = \mathcal{Z} \pi(S_k) \eqsp .  
\end{align}
The important quantities are the weight ratios $w_{k}/w_{k'}$ for $k'\neq k$ for which the normalizing constant $\mathcal{Z}$ simplifies. Therefore, the weights $w_k$ may be defined up to a common multiplicative constant.
Fixing the multiplicative constant, $w_{k}$ can be interpreted as the normalization constant of the target distribution $\pi$ restricted to $S_k$, with un-normalized density given by $x \mapsto \1_{S_k}(x) \gamma(x)$. 
As such, it becomes clear that mode weight estimation is related to the estimation of a normalization constant.

The need to go beyond local MCMC samplers for multi-modal distributions is well established, with annealing-based methods playing a key role (detailed in the next section). However, accurately estimating mode weights, even when mode locations are known, remains an open challenge in high dimensions. This issue parallels free energy computations in statistical mechanics, an active area of research \citep{lelievreFreeEnergyComputations2010,chipotFreeEnergyMethods2023}.

Despite this, new samplers rarely evaluate their accuracy in estimating mode weights. Yet, this criterion complements common choices like integral probability metrics (IPMs) -- such as the Wasserstein distances \citep{peyre2019computational} or the maximum mean discrepancy \citep{gretton2012akernel} -- the estimation of which is complex and typically comes without guarantees like unbiasedness. Mode weights, in contrast, can be easily estimated from the proportion of samples in each mode, offering a more interpretable measure that directly highlights the core challenge of multi-modal sampling.

%% file: 3_samplers.tex
\section{Towards a systematic experiment for evaluating samplers}
\label{sec:3_samplers}

We propose a simple but challenging experimental setting to assess the ability of a given sampler at handling multi-modality by probing systematically two sources of difficulty: dimension and separation of the modes. We illustrate this setting by applying it to a set of common samplers, revealing their strengths and weaknesses in the task of mode weight estimation.

\begin{figure}[t!]
    \centering
    \includegraphics[width=\linewidth]{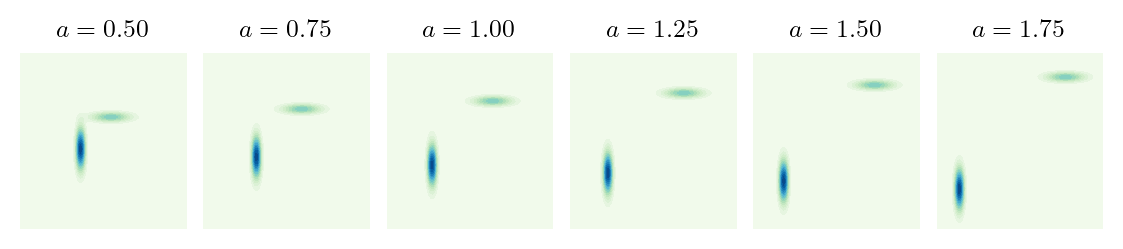}
    \caption{Density of the bi-modal Gaussian mixture used in our experiments ($d = 2$) with increasing values of $a$. As expected, a larger value of $a$ makes the modes further from each other, which reinforces the sampling difficulty.}
    \label{fig:target}
\end{figure}

\subsection{Description of the experimental setting}\label{subsec:def_target}

\paragraph{Multi-modal target distribution.}We consider $\pi$ defined as a $d$-dimensional bi-modal Gaussian mixture with unbalanced weights and covariances with strong components in different directions. More precisely, the mixture components are $q_1(x)=\densityGaussian(x;-a\mathbf{1}_d, \Sigma_1)$, with weight $w=2/3$, and $q_2(x)=\densityGaussian(x;a\mathbf{1}_d, \Sigma_2)$, with weight $1-w=1/3$, where $a\in (0, \infty)$ controls the distance between the local maxima of $\pi$ as showed in \Cref{fig:target}, $\mathbf{1}_d$ is the $d$-dimensional vector with all components equal to $1$, $\Sigma_1$, $\Sigma_2$ are fixed diagonal positive matrices with conditioning number equal to $20$. To make this target even more challenging, we set those covariance matrices such that $q_1$ has marginally large variance in directions where $q_2$ has small variance (and inversely), see \Cref{app:implem} for more details. In this setting, the mode partition of $\pi$ is taken as $S_1=\{x\in \rset^d:q_1(x) >q_2(x)\}$ and $S_2=S_1^c$, such that the induced $w_1$, defined in \eqref{eq:mode_weight}, verifies $w_1 \approx w$.

\begin{figure}[t!]
    \centering
    \includegraphics[width=0.48\linewidth]{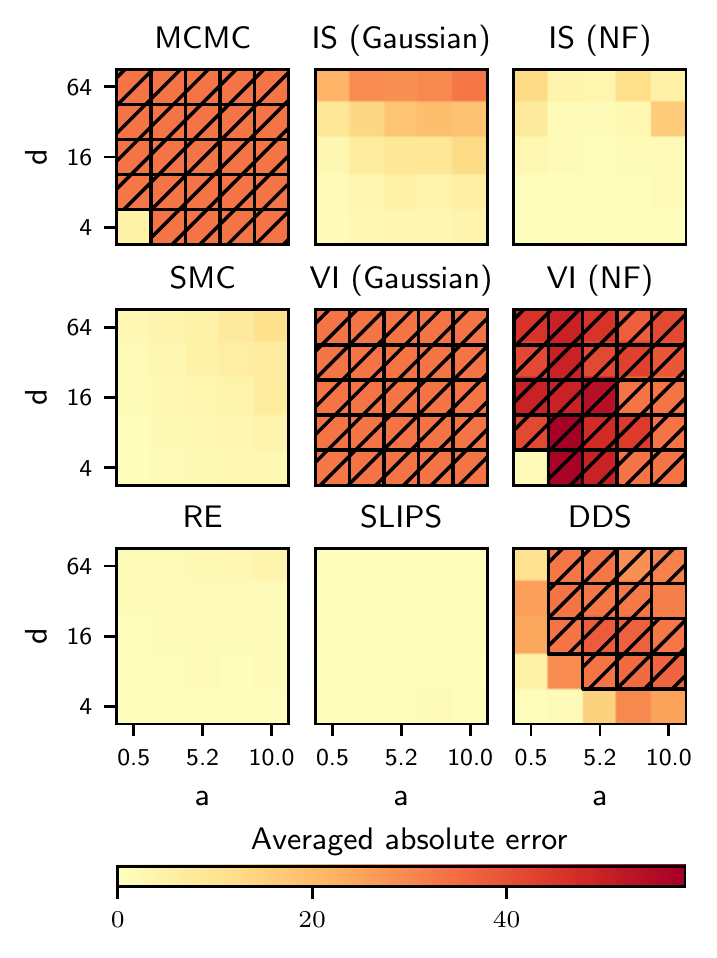}
    \hfill
    \includegraphics[width=0.48\linewidth]{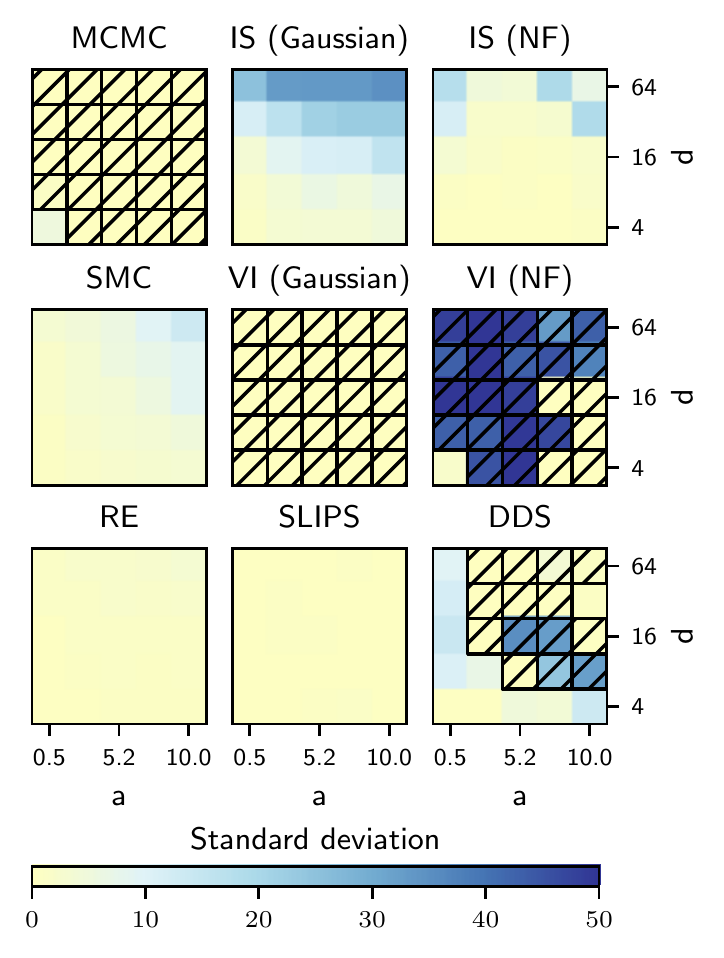}
    \vspace{-0.3cm}
    \caption{Results on mode weight estimation for the bi-modal Gaussian mixture defined in \Cref{subsec:def_target}, when varying hyperparameters $a$ and $d$. For each setting, we aggregate 48 Monte Carlo estimations, each one being computed over 8192 samples. Hashed areas indicate settings with systematic mode collapse in the sampling process. \textbf{(Left):} Averaged absolute error of the estimation with respect to the true mode weight $w\approx 66.7\%$. \textbf{(Right):} Standard deviation of the estimation.}
    \vspace{-0.3cm}
    \label{fig:mode_weight}
\end{figure}

\paragraph{Mode weight metric.}Below, we report the bias and variance for the Monte Carlo estimations of $w_1$,  while varying two  parameters: the distance between the mode centroids, controlled by the parameter $a$, and the dimension $d$. The results are displayed on Figures \ref{fig:mode_weight} and \ref{fig:mode_weight_high_dim} for a selection of sampling methods. We provide, in \Cref{app:implem}, a brief description each method with its  hyperparameters and, below, comment on its performance.  Small bias and variance indicate that the considered sampler is able to recover the mode proportions of our challenging test target both accurately and robustly, and is therefore a promising approach.

\paragraph{General remarks on the methodology.}We emphasize that our numerical results should not be interpreted as a precise or definitive quantitative comparison between samplers for several reasons. First, their computational budgets were only roughly matched, see \Cref{fig:compute_time} in \Cref{app:implem}. 

Secondly, to showcase their best-case performance, their hyperparameters were individually tuned based on access to
ground-truth samples, which would not be available in real-world scenarios, thus making their setting unrealistic for practical deployment.

Additionally, while the experimental target is admittedly a simplified, toy example compared to the complex multi-modal distributions discussed above, we note that most of the considered samplers already fail to perform well on this basic benchmark. This is a key observation: we argue that if a method cannot reliably sample from this controlled bi-modal setting, it is unlikely to succeed in more challenging, high-dimensional applications. Rather than offering fine-grained comparisons, these results are therefore intended to provide insight into the typical behavior and limitations of popular sampling algorithms.

\subsection{Evaluation on various samplers}

\paragraph{Local Markov Chain Monte Carlo.}
MCMC methods produce Markov chains that admit $\pi$ as their stationary measure. Standard designs of transition kernels for log-concave distributions are based on discretizations of either Langevin or Hamiltonian dynamics. However, as already discussed in \Cref{sec:mode_weight}, the resulting sampling algorithms are inefficient when facing multi-modal targets, as moving between distinct modes becomes an unlikely event. 

Testing an implementation of MALA, this behavior is easily evidenced in our experiment. The chains get trapped in one of the modes as soon as there exists a significant log-probability ``barrier" between the modes, i.e., when $a$ or $d$ increases. The zero variance observed in these cases is not to be mistaken as a positive result. As the chains stay trapped in the mode they are initialized in, a setting we refer to as \emph{mode collapsed}, the estimation of $w_1$ is consistently biased in the same manner.

\paragraph{Importance Sampling.}

{Importance Sampling} (IS) builds Monte Carlo estimators as re-weighted averages of samples from a tractable proposal distribution. The variance of IS estimations, related to the degeneracy of the so-called ``importance weights",
depends on the agreement between the target and the proposal distributions and is known to increase exponentially with $d$ \citep{agapiou2017importancesampling}. 

In \Cref{fig:mode_weight}, the bias and variance of the estimator of $w_1$ are reported for two versions of IS. First, using a unimodal Gaussian proposal, the accuracy decreases rapidly along both the mode separation and dimension axes. As either of these parameters increases, the intersection of the typical sets of the target and proposal distribution decreases, leading to more degenerate weights.
Using a normalizing flow has been proposed to adapt the proposal to the target and mitigate the issue \citep{muller2019neural, noe2019boltzmann}. Here, 
the normalizing flow was trained on target samples to serve as a proposal, and indeed mitigates the degradation of performance, in particular with respect to the mode separation. The normalizing flow can learn to focus mass on different modes even if distant. However, the lack of perfect agreements between the target and the proposal starts showing as the dimension increases.

\vspace{-0.2cm}
\paragraph{Variational Inference.}

{Variational Inference} (VI) 
reformulates the sampling problem as an optimization problem over the space of probability distributions. It seeks the closest distribution to the target within a family of tractable parametric distributions, known as the variational family. 
This problem requires a notion of discrepancy between probability distributions, which is 
chosen as the reverse Kullback-Leibler (KL) divergence in most cases. 
For multi-modal target distributions, many works have showed that even for variational family able to represent multiple modes, 
the VI procedure often
leads to mode collapse, due to the mode-seeking behavior of the reverse KL divergence \citep{mackay2001local, jerfelVariationalRefinementImportance2021a, blessing2024beyond, soletskyi2024theoreticalperspectivemodecollapse}.

In our experiment, we consider two families of variational distributions, 
unimodal Gaussian distributions and normalizing flows \cite{rezendeVariationalInferenceNormalizing2015}. In the Gaussian case, we systematically observe a mode collapsed estimation, with the same consistent bias as in the MCMC setting, resulting in a zero-variance outcome.
While using normalizing flows remediates the issue when the modes are almost connected and the dimension is low (lowest values of $a$ and $d$), mode collapse is systematic beyond this setting.

\paragraph{Annealed sampling methods.}
Annealed samplers were developed to address the challenges posed by multi-modal distributions. The core idea is to introduce a sequence of distributions that gradually transition from an easy-to-sample distribution to the target distribution. These samplers leverage this sequence by breaking the original sampling problem into multiple, more manageable subproblems. In practice, this is often achieved through a geometric interpolation of densities.

\subparagraph{Sequential Annealed Sampling.}
A naive annealing approach consists in performing recursive MCMC sampling of the bridging distributions. Beginning with the simplest distribution and sequentially moving towards the target, each intermediate distributions is targeted by a standard MCMC sampler, that uses samples obtained for the preceding distribution as a warm-start.
The Sequential Monte Carlo (SMC) algorithm \citep{del2006sequential} improves upon this approach by reweighting warm-start samples at certain steps using importance weights updated throughout the process. While SMC significantly outperforms the earlier samplers in handling multi-modality, it has been shown to degrade in high dimensions. Similar to Importance Sampling (IS), its resampling weights suffer from degeneracy \citep{li2014fight}, a problem that can be exacerbated by the default geometric interpolation path \citep{mate2023learning_bis}. Increasing the number of intermediate steps can help mitigate this issue in higher dimensions but comes at a significant computational cost.

\begin{figure}[t!]
    \centering
    \includegraphics[width=0.48\linewidth]{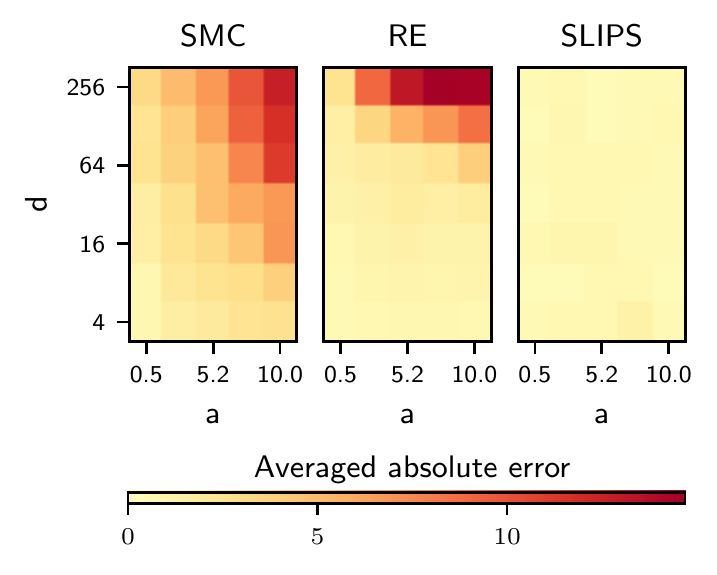}
    \hfill
    \includegraphics[width=0.48\linewidth]{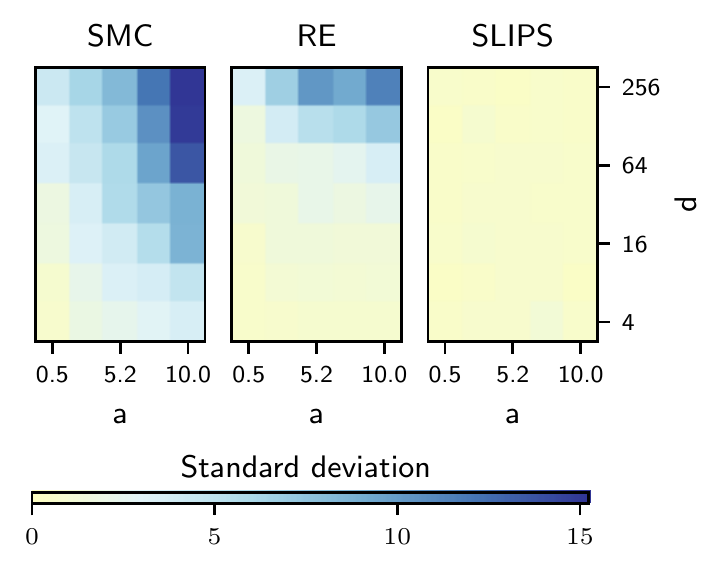}
    \vspace{-0.3cm}
    \caption{
    Same as \Cref{fig:mode_weight} for the best performing SMC, RE and SLIPS reaching higher-dimensions. For sake of pedagogy, the color scale is different from above (a consistent color scale can be found in \Cref{fig:mode_weight_high_dim_orig_scale} of \Cref{app:implem}).
    \textbf{(Left):} Averaged absolute error of the estimation with respect to the true mode weight $w\approx 66.7\%$. \textbf{(Right):} Standard deviation of the estimation.}
    \vspace{-0.3cm}
    \label{fig:mode_weight_high_dim}
\end{figure}

For our experiment, we leverage an optimal sequence of intermediate distributions based on the recommendations of \cite{chopin2020introduction}, placing SMC in its most favorable setting. In \Cref{fig:mode_weight}, we observe that SMC is perfectly able to recover the true mode proportion with high accuracy and robustness, confirming the superiority of annealing schemes over standard MCMC methods for multi-modal distributions. However, as expected, SMC performance deteriorates as soon as both of the mode separation and the dimension increase together as shown in \Cref{fig:mode_weight_high_dim}.

\subparagraph{Parallel annealed sampling.}

A second approach amounts to perform MCMC sampling of each intermediate distribution in parallel, and to exchange states from the generated adjacent chains based on a Metropolis-Hastings acceptance step. This algorithm is often referred to as {Replica Exchange} (RE) \citep{swendsen1986replica} or {Parallel Tempering} (PT) \citep{geyer1991markov, hukushima1996exchange}. Similarly to sequential annealing, this approach is known to suffer from high dimension, due to the degeneracy of the acceptance probabilities.

Here, we test an implementation of RE with a number of intermediate distributions that is fixed over the choices of $a$ and $d$ \footnote{We explored various types of discretization schedules for the intermediate distributions and observed that a logarithmic spacing was the best suited for this target.}. Although RE accurately recovers the true mode proportion in the settings from \Cref{fig:mode_weight}, its performance declines with increasing mode separation and dimension, as showed in \Cref{fig:mode_weight_high_dim}. This degradation appears slower than SMC, but note the adaptive SMC converged in a computational budget 20 times faster than the one considered for RE.

\paragraph{Diffusion-based sampling methods.}
Building on classical sequential annealing, the recently introduced diffusion-based sampling framework generates samples from the target distribution by following a Stochastic Differential Equation (SDE) that transitions from an easy-to-sample distribution to the target via a predefined stochastic interpolant process.
This approach is inspired by the success of score-based diffusion models in generative modeling \citep{song2020score, albergo2023stochastic}, where the goal is to generate samples without access to an explicit density. In practice, the drift term of this generative SDE is intractable and must be estimated, typically using Monte Carlo methods or variational approximations.

\subparagraph{Monte-Carlo diffusion-based sampling.}
Leveraging the fact that the intractable drift can be expressed as a conditional expectation, several recent works have proposed estimating it on-the-fly—i.e., while running the generative SDE—using unparameterized Monte Carlo methods \citep{huang2024reverse,huang2024fastersamplingisoperimetrydiffusionbased,he2024zerothorder}.
A related approach, Stochastic Localization via Iterative Posterior Sampling (SLIPS) \citep{grenioux2024stochastic}, estimates the drift using standard MCMC techniques. While these methods have demonstrated strong empirical performance, they require careful hyperparameter tuning which becomes particularly challenging when no ground truth samples are available for their calibration, which is often the case in practice. This represents a critical limitation for their real-world applicability.

With this informed tuning, SLIPS performs very well at the mode weight estimation task as presented in \Cref{fig:mode_weight} and even outperforms SMC and RE in higher dimension, as displayed in \Cref{fig:mode_weight_high_dim}. This result may be explained by the fact that the sequence of intermediate densities induced by the diffusion path (which are also bi-modal) {preserve} the mode weights of the target distribution. This is contrary to the geometric interpolation paths \citep{mate2023learning_bis} and avoids the use of advanced resampling schemes that notoriously fail when $a$ or $d$ increases.

\subparagraph{Variational diffusion-based sampling.}
Another line of research has proposed to learn the drift of the generative SDE with a neural network, by solving a Variational Inference problem, this time defined on the space of path measures \citep{zhang2021path, berner2022optimal, vargas2023transport,chen2024sequential}. In this case, one aims to minimize a discrepancy between the generative SDE and a variational family of neural SDEs, i.e., SDEs whose drift is a neural network. In our experiment, we considered the Denoising Diffusion Sampler (DDS) \citep{vargas2023denoising}.
The interest of such parameterized methods lies in their ability to amortize inference; once the drift is learned, samples can be obtained by solving the corresponding neural SDE. 
Their performance were found to severely depends on the values of their hyperaparameters \cite{noble2024learned}. 

In \Cref{fig:mode_weight}, we display results of DDS, where its hyperparameters were tuned favorably with respect to $\pi$, see \Cref{app:implem}.
Despite relative good performance in the most simple multi-modal settings, we still observe that it suffers from the joint increase of $a$ and $d$. 
The difference with the Monte-Carlo diffusion-based sampler (SLIPS) may be explained by a manifestation of mode collapse in this extended VI problem.

%% file: 4_conclusion.tex
\section{Perspectives}
We have demonstrated that evaluating a sampler’s ability to recover mode weights accurately provides an interpretable benchmark for designing samplers for multi-modal distributions. Gradually increasing both dimensionality and mode separation in a controlled experiment reveals the breaking points of different samplers and helps identify the most promising approaches.
Designed to handle multi-modal distributions, traditional annealing-based samplers, such as SMC and RE, are generally more robust than direct sampling methods. Recently introduced diffusion-based samplers also show promise, but their precise tuning and scalability remain open challenges.

While we focused on dimension and mode separation, the number of modes also adds to the challenge of sampling. The proposed experimental setting can be straightforwardly extended in this third axis of difficulty. In this case, the mode weight criterion translates into a comparison of the empirical histogram of visited modes with the ground-truth components' weights (see e.g., \cite{grenioux2023onsampling}). We also note that the proposed framework does not include heavy-tails and complex local geometries, and it would be an interesting direction to define a systematic benchmark including also these features.

%% file: 0_appendix.tex
\section{Details on considered samplers}

In the following, we present a brief description of the samplers presented in \Cref{sec:3_samplers}, and used in our numeric evaluation. Throughout this section, we will consider an arbitrary test function $f:\rset^d \to \rset$, for which we would like to evaluate the expectation $\mathbb{E}_\pi[f]:=\int_{\rset^d}f(x)\rmd\pi(x)$, and will explain how to compute Monte Carlo estimations $\mathbb{E}_\pi[f]$ for each sampler. Note that in our numerical experiment, we will take $f=\1_{S_1}$. Below, $p^{\text{init}}$ will refer to an easy-to-sample distribution, with tractable density and normalization constant (basically, a Gaussian distribution).

\paragraph{MALA sampler.} This algorithm is an extension of the largely used Unadjusted Langevin Algorithm (ULA) \citep{roberts1996exponential}, which builds a Markov chain $(X^\ell)_{\ell=0}^L$ with $(L+1)$ states ($L \geq 1)$, defined by $X^0 \sim p^{\text{init}}$ and for any $\ell \in \{0, \ldots, L-1\}$,
\begin{align}
    X^{\ell+1} = X^\ell + \lambda \nabla \log \gamma(X^\ell) + \sqrt{2 \lambda} Z^\ell \eqsp, \eqsp Z^\ell \sim \densityGaussian(0, \Idd) \eqsp,
\end{align}
which reduces to a time discretization of the Langevin diffusion that admits $\pi$ as invariant distribution. Building upon this recursion, MALA \citep{roberts1996exponential} addresses the time discretization bias by adding a Metropolis-Hastings (MH) acceptance/rejection step at each iteration $\ell$ based on the acceptance probability
\begin{align}
    \alpha(x,x')=\min \left(1, \frac{\gamma(x')\densityGaussian(x; x'+ \lambda \nabla \log \gamma(x'), 2 \lambda \Idd)}{\gamma(x)\densityGaussian(x'; x+ \lambda \nabla \log \gamma(x), 2 \lambda \Idd)}\right) \eqsp .
\end{align}
In this case, the corresponding Monte Carlo estimation of $\mathbb{E}_\pi[f]$ will be given by $L^{-1}\sum_{\ell=1}^L f(X^\ell)$.

\paragraph{Importance Sampling.} Given an arbitrary proposal probability distribution $\nu$ with tractable density $q$ known up to a normalizing constant such that $\pi \ll \nu$, Importance Sampling (IS) relies on the following equality
\begin{equation}\label{eq:is}
     \mathbb{E}_\pi[f] = \mathbb{E}_\nu\left[\frac{\rmd \pi}{\rmd \nu}f\right] = \frac{1}{\mathcal{Z}}\mathbb{E}_\nu\left[\frac{\gamma}{q}f\right] \eqsp .
 \end{equation}
Hence, the IS Monte Carlo estimation of $\mathbb{E}_\pi[f]$ is given by the weighted mean $N^{-1}\sum_{n=1}^N w_n(X^n) f(X^n)$, where $(X^n)_{n=1}^N$ are samples from $\nu$ and $\{w_n(X^n)\}_{n=1}^N$ are self-normalized importance weights defined by 
  \begin{align}
     w_n(X^n) = \frac{\tilde{w}_n(X^n)}{\sum_{m=1}^N \tilde{w}_m(X^m)} \eqsp , \eqsp \tilde{w}_n(X^n) = \frac{\gamma(X^n)}{q(X^n)} \eqsp .
 \end{align}
In the special case where the proposal is defined by a normalizing flow $\mathrm{T}_\theta:\rset^d \to \rset^d$ 
with base distribution $p^{\text{init}}$, the corresponding parametric density is given by
\begin{align}
    q_\theta(x) = p^{\text{init}}(\mathrm{T}_\theta^{-1}(x)) \abs{J_{\mathrm{T}_\theta^{-1}}\left(x\right)} \eqsp .
\end{align}
where $J_{\mathrm{T}_\theta^{-1}}$ denotes the Jacobian of $\mathrm{T}_\theta^{-1}$.

\paragraph{Variational Inference.} Given a family of parametric distributions $\{\nu_\theta\}_{\theta \in \Theta}$ such that for any $\theta\in \Theta$, (i) $\nu_\theta$ admits a tractable density $q^\theta$ known up to a normalizing constant and (ii) $\nu_\theta$ is easy to sample from, VI aims to solve the following optimization problem
\begin{align}
    \theta^\star \in \arg\min_{\theta\in \Theta} \operatorname{D}(\nu_\theta|\pi) \eqsp ,
\end{align}
where $\operatorname{D}$ is a divergence\footnote{Commonly, a divergence $D$ verifies $\operatorname{D}(\nu \mid \pi) \geq 0$ and $\operatorname{D}(\nu \mid \pi) = 0$ if and only if $\nu=\pi$.} operating on the space of probability distributions. In most cases, $\operatorname{D}$ is defined as the reverse Kullback-Leibler divergence, i.e., we have $\operatorname{D}(\nu_\theta|\pi)= \operatorname{KL}(\nu_\theta|\pi)=\int_{\rset^d}\log(q_\theta(x)/\gamma(x)) \rmd \nu_\theta(x)$, where the equality holds up to constants independent from $\theta$. In practice, this objective function is minimized by performing Stochastic Gradient Descent (SGD), based on samples from the variational distribution being trained. Assuming that $\hat{\theta}$ is the output of this numerical procedure, the corresponding VI Monte Carlo estimation of $\mathbb{E}_\pi[f]$ is given by $N^{-1}\sum_{n=1}^N f(X_{\hat{\theta}}^n)$, where $(X_{\hat{\theta}}^n)_{n=1}^N$ are samples from $\nu_{\hat{\theta}}$.

\paragraph{Annealed sampling.} Given $K\geq 1$ the number of intermediate steps, the standard geometric density interpolation defines a sequence of target densities $\{p_k\}_{k=0}^K$ up to a normalizing constant by $p_k \propto (p^{\text{init}})^{1-\beta_k}\gamma^{\beta_k}$, where $k\in\{0, \hdots, K\}$ and $\{\beta\}_{k=0}^K$ is an increasing sequence with $\beta_0=0$ and $\beta_{K}=1$. In the literature, $\{\beta\}_{k=0}^K$ is often referred to as the tempering schedule.

\paragraph{Sequential Monte Carlo.} Given $N\geq 1$, SMC generates $N$ chains of size $K+1$, referred to as particles, such that the $k$-th elements of all chains are expected to be samples from $p_k$. To do so, SMC relies \emph{locally} on MALA steps performed independently on each chain and \emph{globally} on resampling steps based on accumulated importance weights, as defined next. 

Let $k\in \{0, \hdots, K-1\}$, $(\tilde{w}^n_{k})_{n=1}^N$ be unnormalized importance weights accumulated on each chain up to step $k$ ($\tilde{w}^n_{k}=1/N$ if $k=0$), and $(x^n_{k})_{n=1}^N$ be samples from $p_k$ obtained independently as the last samples of $N$ parallel MALA runs (or exactly obtained if $k=0$). Then, the SMC importance weights designed to sample from $p_{k+1}$, denoted by $(\tilde{w}^n_{k+1})_{n=1}^N$, and their renormalized versions, denoted by $(w^n_{k+1})_{n=1}^N$, are defined as
\begin{align}
    \tilde{w}^n_{k+1} = \frac{p_{k+1}(x^n_{k})}{ p_{k}(x^n_{k})} \tilde{w}^n_k, \quad w^n_{k+1} = \frac{\tilde{w}^n_{k+1}}{\sum_{j=1}^N \tilde{w}^j_{k+1}}\eqsp.
\end{align}
The SMC resampling step then consists in obtaining $N$ new samples $(y^n_{k+1})_{n=1}^N$ by random selection among the original samples $(x^n_{k})_{n=1}^N$ via a multinomial distribution (with replacement) defined by the normalized weights $(w^n_{k+1})_{n=1}^N$. These novel samples will then be used as starting points for running MALA on the target $p_{k+1}$. In our implementation, we follow the recommendation of \cite{chopin2020introduction} and build the sequence $\{\beta_k\}_{k=0}^K$ on-the-fly. This is done by solving, at each step $k$, the following optimization problem
\begin{align}
    &\delta^\star_k = \min\{ \delta \in [0, 1-\beta_k] : \operatorname{ESS}(\delta) = \operatorname{ESS}_{\text{min}}\} \eqsp ,\\
    &\text{where }\operatorname{ESS}(\delta) = \left[\left(\sum_{n=1}^N \tilde{w}^n_{\delta, k}\right)^2/\sum_{n=1}^N(\tilde{w}^n_{\delta, k})^2\right] \quad  \tilde{w}^n_{\delta, k} = \frac{(p^{\text{init}})^{1-(\beta_k+\delta)}(x^n_{k})\gamma^{\beta_k + \delta}(x^n_{k})}{ p_{k}(x^n_{k})}\eqsp,
\end{align}
and $\operatorname{ESS}_{\text{min}}=\alpha N$, with $\alpha\in (0,1)$ being a user-defined threshold. Then, the schedule update is given by $\beta_{k+1} = \beta_k + \delta^\star_k$. By doing so, this procedure enables to prevent the degeneracy of the particles at step $k+1$ (which occurs when the corresponding ESS is too low). In practice, we minimize this quantity using Brent's method.
In the end, the SMC estimator of $\mathbb{E}_\pi[f]$ is given by the weighted mean $N^{-1}\sum_{n=1}^N w^n_K f(X_K^n)$, where $(X_K^n)_{n=1}^N$ are the last MALA samples obtained at step $K$.

\paragraph{Replica Exchange.} The RE algorithm aims at sampling from the annealed distributions $\{p_k\}_{k=0}^K$ with $K+1$ parallel Markov chains, i.e., the $k$-th chain targets $p_k$. For the sake of pedagogy, we assume here that $K$ is even. 

At each level $k$, RE sampling is done via a local MCMC sampler, such as MALA, which is expected to have poor performance for large $k$ (as $p_k$ will be highly multi-modal). To alleviate this issue, every $L$ MALA steps, RE randomly performs a \emph{swapping} between Markov chains. More specifically, each consecutive pair $(k, k+1)$ of Markov chains (i.e., either $(k, k+1)\in \{(0,1),\ldots,(K-2,K-1)\}$ or $(k, k+1)\in \{(1,2),\ldots,(K-1,K)\}$), with respective current states $X_k$ and $X_{k+1}$, is swapped with probability
\begin{align}
    p_{k,k+1} = \min\left(1, \frac{p_k(X_{k+1}) p_{k+1}(X_k)}{p_{k}(X_k) p_{k+1}(X_{k+1})}\right)\eqsp.
\end{align}
After running the Markov chains for $N$ steps, we obtain a Monte Carlo estimation of $\mathbb{E}_\pi[f]$ by considering $N^{-1}\sum_{n=1}^N f(X_K^n)$, where $(X_K^n)_{n=1}^N$ are the visited states of the $K$-th chain (corresponding to the target distribution).

\newpage
\paragraph{SLIPS.} \cite{grenioux2024stochastic} proposed a diffusion-based sampling method extending the stochastic localization framework from \cite{montanari2023sampling}, which can be seen as an analog of denoising diffusion models. More precisely, given a horizon time $T>0$, they consider a stochastic process $(Y_t)_{t\in [0,T]}$, called observation process, defined by
\begin{align}
    Y_t = \alpha(t) X + \sigma W_t \eqsp , \eqsp X \sim \pi \eqsp ,
\end{align}
where $\sigma \in(0, \infty)$, $\alpha:[0,T)\to [0, \infty)$ is an increasing function of time verifying $\alpha(0)=0$ and $\alpha(t)/\sqrt{t}\to \infty$ as  $t\to T$, and $(W_t)_{t\geq 0}$ is a $d$-dimensional Brownian motion. A particular choice of interest is given by $\alpha(t)=\sqrt{t/(1-t)}$, referred to as $\operatorname{Geom}(1,1)$, with $T=1$.

By definition of $\alpha$, this observation process \emph{localizes} on the target distribution as time increases since $Y_t/\alpha(t) \to X$ as $t\to T$. Hence, if we were able to simulate the stochastic process $(Y_t)_{t\in [0,T]}$ up to time $T$ (which is infeasible a priori since we do not have access to ground truth samples), we would then obtain approximate samples from $\pi$. To do so, the authors propose to simulate the Markov process induced by the following SDE, which is proved to have the same marginals as the observation process, denoted by $(p_t)_{t\in [0,T]}$,
\begin{align} \label{eq:slips}
    \rmd Y_t = \dot{\alpha} (t) u_t(Y_t) \rmd t + \sigma \rmd B_t \eqsp , Y_0=0 \eqsp ,
\end{align}
where $(B_t)_{t\geq 0}$ is another $d$-dimensional Brownian motion and $u_t(y_t)=\mathbb{E}[X|Y_t=y_t]$ is the conditional expectation of the posterior distribution whose unnnormalized density is given by $q_t(x|y_t): x \mapsto \densityGaussian(y_t; \alpha(t)x, \sigma^2t\Idd)\gamma(x)$. This drift being intractable, the authors propose to estimate it on-the-fly, i.e., while running a discretized version of the SDE \eqref{eq:slips}, using MCMC methods to target the posterior $q_t$. They justify their method by highlighting the phenomenon \emph{duality of log-concavity}, that is, while $p_t$ gets increasingly multi-modal, $q_t$ gets increasingly more log-concave and therefore, easier to sample from with standard MCMC techniques. Overall, the success of this method relies on tuning a critical hyperparameter $t_0\in (0,T)$, which denotes the effective starting time of the generative SDE, such that both $p_{t_0}$ and $q_{t_0}$ are approximately log-concave. While some heuristics have been proposed by the authors to solve this issue, they may be not effective on arbitrary multi-modal distributions. For this algorithm, given a time-step $t_1$ close to $T$, the corresponding Monte Carlo estimation of $\mathbb{E}_\pi[f]$ is given by $N^{-1}\sum_{n=1}^N f(Y_{t_1}^n/\alpha(t_1))$, where $(Y_{t_1}^n)_{n=1}^N$ are obtained by running independently $N$ times the approximated version of \eqref{eq:slips}, up to time $t_1$.

\paragraph{DDS.} This parameterized diffusion-based sampling method can be seen as the analog of the Variance-Preserving (VP) score-based generative model previously introduced by \citep{song2020score}, in the setting where ground truth samples are not available. More precisely, given a time horizon $T>0$ and $\sigma\in (0, \infty)$, DDS aims to sample from a \emph{denoising diffusion process} $(Y_t)_{t\in [0,T]}$ defined as the time-reversal of the Ornstein-Uhlenbeck process $(X_t)_{t\in [0,T]}$ induced by the following so-called \emph{noising} SDE
\begin{align}
    \rmd X_t =-\frac{\beta(t)}{2}X_t \rmd t + \sigma \sqrt{\beta(t)} \rmd W_t \eqsp , X_0 \sim \pi \eqsp,
\end{align}
where $(W_t)_{t\geq 0}$ is a $d$-dimensional Brownian motion and $\beta:[0,T]\to (0,\infty)$ is a noising schedule chosen such that it holds approximately $X_T\sim \densityGaussian(0, \sigma^2 \Idd)$. In particular, it holds that $Y_t=X_{T-t}$ in the distribution sense, and, under mild assumptions on $\pi$ and $\beta$, $(Y_t)_{t\in [0,T]}$ is solution to the following non-linear SDE
\begin{align}
    \rmd Y_t =\frac{\beta(T-t)}{2}Y_t \rmd t + \sigma^2\beta(T-t) \nabla \log p_{T-t}(Y_t) \rmd t + \sigma \sqrt{\beta(t)} \rmd B_t \eqsp , Y_0 \sim \densityGaussian(0, \sigma^2 \Idd) \eqsp,
\end{align}
where $(B_t)_{t\geq 0}$ is another $d$-dimensional Brownian motion and $p_t$ denotes the density of the marginal distribution of $X_t$, which is intractable. Hence, in the same spirit as SLIPS, if we were able to simulate the stochastic process $(Y_t)_{t\in [0,T]}$ up to time $T$ (which is infeasible a priori since the scores $\nabla \log p_{t}$ are intractable), we would then obtain approximate samples from $\pi$. To do so, \cite{vargas2023denoising} propose to learn the scores by solving the following Variational Inference problem, which does not require access to ground truth samples, 
\begin{align}
    \theta^\star \in \arg \min_{\theta \in \Theta}\operatorname{KL}(\mathbb{P}^\theta|\mathbb{P}) \eqsp ,
\end{align}
where $\mathbb{P}$ is the path measure associated to the true reverse process, and for any $\theta \in \Theta$, $\mathbb{P}^\theta$ is the path measure associated to the neural SDE
\begin{align} \label{eq:neural_SDE}
    \rmd Y_t =\frac{\beta(T-t)}{2}Y_t \rmd t + \sigma^2\beta(T-t) s_\theta(T-t,Y_t) \rmd t + \sigma \sqrt{\beta(t)} \rmd B_t \eqsp , Y_0 \sim \densityGaussian(0, \sigma^2 \Idd) \eqsp,
\end{align}
where $s_\theta:[0,T] \times \rset^d \to \rset^d$ is a neural network learnt such that $s_\theta(t,\cdot)\approx\nabla \log p_t$. To obtain a tractable objective function, \cite{vargas2023denoising} propose to parameterize $s_\theta$ via a linear term as $s_\theta(t,x)=-\frac{x}{\sigma^2} + u_\theta(t,x)$, where $u_\theta$ can be interpreted as a guidance. After time discretization, the optimization procedure to learn $\theta^\star$ amounts to perform VI on joint densities. Note that the reverse KL divergence may be replaced by the Log-Variance divergence for increased performance \cite{richter2023improved}. Assuming that $\hat\theta$ is the output of this numerical procedure, the corresponding DDS Monte Carlo estimation of $\mathbb{E}_\pi[f]$ is given by $N^{-1}\sum_{n=1}^N f(Y_{\hat\theta,T}^n)$ where $(Y_{\hat\theta,T}^n)_{n=1}^N$ are obtained by running independently $N$ times the neural SDE \eqref{eq:neural_SDE} with control term $s_{\hat \theta}$, up to time $T$.

\section{Implementation details}
\label{app:implem}

\paragraph{Details on the multi-modal target distribution.} In this paper, we consider the Gaussian mixture with density defined by
\begin{align}\label{eq:target-density}
    \gamma : x \in \rset^d \mapsto \frac{2}{3}\overbrace{\densityGaussian(x;-a\mathbf{1}_d, \Sigma_1)}^{q_1(x)} + \frac{1}{3}\underbrace{ \densityGaussian(x;a\mathbf{1}_d, \Sigma_2)}_{q_2(x)} \eqsp ,
\end{align}
where $\Sigma_1\in \rset^{d\times d}$ and $\Sigma_2\in \rset^{d\times d}$ are both diagonal matrices respectively defined by $(\Sigma_1)_{i,i}=\frac{i}{d}\sigma^2_{\max} + \frac{d-i}{d}\sigma^2_{\min}$ for $i\in \{1, \hdots, d\}$ and $(\Sigma_2)_{i,i}=(\Sigma_1)_{d-i,d-i}$ for $\sigma^2_{\max}=0.2$ and $\sigma^2_{\min}=0.01$. In our numerics, we considered the following range of hyperparameters: $a\in \{0.5, 2.875, 5.25, 7.625, 10\}$ and $d\in \{4,8,16,32,64\}$ (and the additional settings $d\in \{128,256\}$ for SMC, RE and SLIPS). Below, we will consider the Gaussian approximation of $\pi$ defined by $\densityGaussian(\mathbf{m}_\pi, \Sigma_\pi)$, where $\mathbf{m}_\pi$ is the (tractable) mean of $\pi$ and $\Sigma_\pi$ is the diagonal positive matrix such that for any $i\in \{1, \hdots, d\}$, $\operatorname{Diag}(\Sigma_\pi)_{i}$ is the marginal variance of $\pi$ on the $i$-th axis (which is also tractable). 

\paragraph{Setting of MALA.} To obtain one Monte Carlo estimation, we consider 32 independent chains randomly initialized in the strongest mode of $\pi$ (i.e., in $S_1$). While running the Markov chains, we automatically adapt the step-size $\lambda$ (initialized at $10^{-4}$) by targeting a specific MH acceptance ratio fixed to $0.75$. For each chain, we first run 4096 warm-up steps, and then run MALA for 8192 steps to obtain as many samples. This procedure produces $8192 \times 32$ samples which will be used to compute the mode weight estimation. We repeat this process $48$ times to obtain aggregate results.

\paragraph{Setting of IS.} To ensure an ideal setting for IS estimation, we set the proposal $\nu$ to be closely adapted to $\pi$, while avoiding giving access to the knowledge of the true mode proportions. This is done by minimizing the forward Kullback-Leibler divergence between the \emph{equilibrated} version of $\pi$, defined by the mixture $x\mapsto \frac{1}{2}q_1(x) + \frac{1}{2}q_2(x)$, and the chosen parametric family of distributions (namely, Gaussian distributions or normalizing flows). Note that sampling from the equilibrated mixture is highly similar to directly sampling from the target distribution with an MCMC procedure, where the location of the modes is given.
Regarding the parametric family used:
\begin{itemize}[wide,label=$\bullet$]
    \item \underline{Gaussian distributions}: we compute maximum likelihood estimators of the ideal parameters, based on 16384 samples of the equilibrated target;
    \item \underline{Normalizing flows (NFs)}: we restrict our choice of parametrization to the RealNVP architecture \citep{dinh2017density}, which is composed of 32 affine coupling blocks with scale and shifts parametrized as Multi-Layer Perceptrons (MLPs) with 3 hidden layers of size 64. The base of the NF is a standard centered Gaussian distribution at temperature 0.5. The NF is trained on a dataset of 16384 samples for 100 epochs with a batch size of 1024 samples. We follow the implementation from \cite{Stimper2023}.
\end{itemize}
Once training the proposal is done, we compute for each setting 48 IS estimators of the mode weight, using 8192 particles each time. 

\paragraph{Setting of VI.} We consider two different variational families :
\begin{itemize}[wide,label=$\bullet$]
    \item \underline{Gaussian distributions}: we restrict our choice to distributions with diagonal covariances. The initial parameters of the VI procedure are taken as the exact mean of $\pi$ and the diagonal of the exact covariance of $\pi$. In practice, we parameterize the log-variances instead of the variances;
    \item  \underline{Normalizing flows (NFs)}: we restrict our choice of parametrization to the RealNVP architecture \citep{dinh2017density}, which is composed of 32 affine coupling blocks with scale and shifts parametrized as MLPs with 3 hidden layers of size 64. The base of the NF is a standard centered Gaussian distribution. We follow the implementation from \cite{Stimper2023}.
\end{itemize}
In both cases, we minimize the reverse Kullback-Leibler divergence for 2048 SGD iterations with a batch size of 2048. Once the optimization is done, we sample 8192 from the obtained optimal parametric distribution to compute the mode weight estimation (this is done 48 times).

\begin{figure}[t!]
    \centering
    \includegraphics[width=0.48\linewidth]{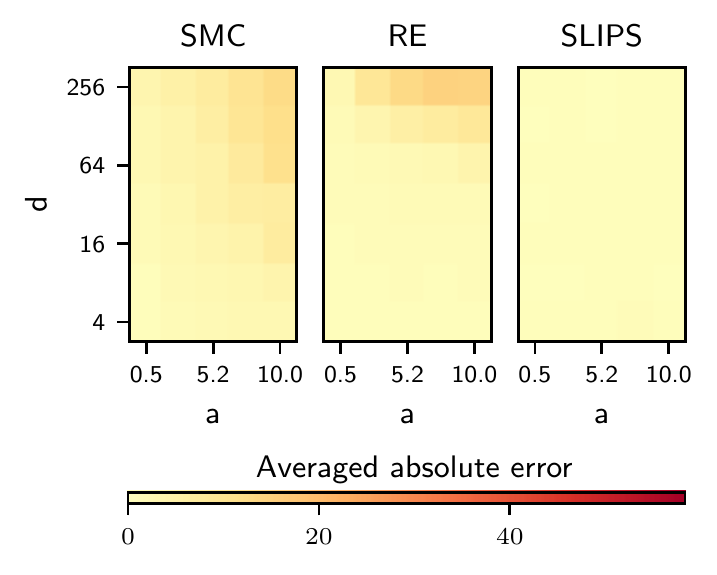}
    \hfill
    \includegraphics[width=0.48\linewidth]{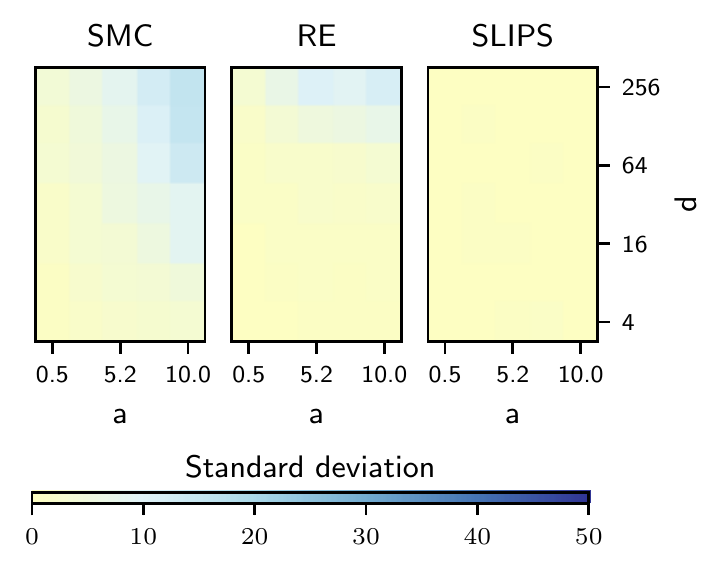}
    \caption{Same as \Cref{fig:mode_weight} for the best performing SMC, RE and SLIPS in higher-dimensions. Note that the color scale is the same as in \Cref{fig:mode_weight}.
    \textbf{(Left):} Averaged absolute error of the estimation with respect to the true mode weight $w\approx 66.7\%$. \textbf{(Right):} Standard deviation of the estimation.}
    \label{fig:mode_weight_high_dim_orig_scale}
\end{figure}

\paragraph{Setting of SMC.} The base distribution $p^{\text{init}}$ is given by $\densityGaussian(\mathbf{m}_\pi, \Sigma_\pi)$.
We set the maximum number of intermediate steps as 512 and $\operatorname{ESS}_{\text{min}} = N/2$ (i.e., $\alpha=1/2$) as per \cite{chopin2020introduction}.
Locally, we perform MALA sampling, with adaptive step-size (initialized at $10^{-2}$), for $96$ steps. To ensure efficiency over the sequential procedure, we initialize the step-size of the $(k+1)$-th sampler with the last value of the step-size from the $k$-th sampler. Overall, we run SMC 48 times, with $N=8192$ particles each time to compute the mode weight.

\paragraph{Setting of RE.} The base distribution $p^{\text{init}}$ is given by $\densityGaussian(\mathbf{m}_\pi, \Sigma_\pi)$. We consider $K=64$ intermediate steps and the schedule $\{\beta_k\}_{k=0}^K$ to be logarithmically spaced between $\varepsilon = 10^{-5}$ and $1$, i.e., $\beta_k=1- \varepsilon^{(k/K)}$ for $k\in \{0,\hdots, K\}$. In our experiments, we observed that this particular choice of schedule (unlike the linear schedule) has benefits to prevent the issue of mode switching in geometric density interpolations \citep{mate2023learning_bis}. Such as in SMC, we locally perform MALA sampling, with adaptive step-size (initialized at $10^{-2}$). In particular, we enable the swapping every 8 MCMC steps. Overall, we perform 16384 warm-up steps at each annealing level and then 32768 steps to generate the samples, for 16 parallel instances of the algorithm. In the end, we perform a thinning by only selecting one sample every 8 steps in the $K$-th chain (corresponding to the target distribution), and randomly sub-sample 8192 samples out of all the obtained samples to compute the mode weight estimation. We repeat this procedure 48 times.

\paragraph{Setting of SLIPS.} We run the SLIPS algorithm with the same setting as in the original paper, see \cite{grenioux2024stochastic}, but we rather use 512 discretization timesteps. Moreover, the Gaussian initialization procedure of SLIPS at time $t_0$ is preserved, but with the additional information of the true mean of the target distribution. As in the original paper, we tune the $t_0$ hyperparameter by leveraging access to ground truth samples. Overall, to obtain the Monte Carlo estimations, we run SLIPS $48 \times 8192$ times.

\paragraph{Setting of DDS.} We run the DDS sampler using the code from \cite{berner2022optimal}, using 100 integration steps uniformly distributed between $0$ and $T$. Following the guidelines from \cite{vargas2023denoising}, we use a target-informed parametrization of the neural network $u^\theta$. Moreover, the hyperparameter $\sigma$ is set according to the recommendations of \cite{noble2024learned}
\begin{equation}
    \sigma^2 = \frac{1}{d} \left(m^\top m + \sum_{i=1}^d \Gamma_i^2\right)
\end{equation}
where $m=\mathbb{E}_\pi[X]=a(2-w)\mathbf{1}_d$ and for $i\in \{1, \hdots, d\}$,
\begin{align}
    \Gamma_i^2 = w \{ (\Sigma_1)_{ii} +(a + m_i)^2\}  + (1-w)\{ (\Sigma_2)_{ii} +(a - m_i)^2\} \eqsp.
\end{align}
Besides this, we train $16$ times the DDS log-variance objective for 4096 gradient steps using a batch size of 1024 samples, and for each training, and compute, for each training, 3 Monte Carlo estimations of the mode weight based on 8192 generated samples (hence, overall 48 MC estimations).

\paragraph{Computational footprint.}

\begin{figure}[t!]
    \centering
    \includegraphics[width=0.5\linewidth]{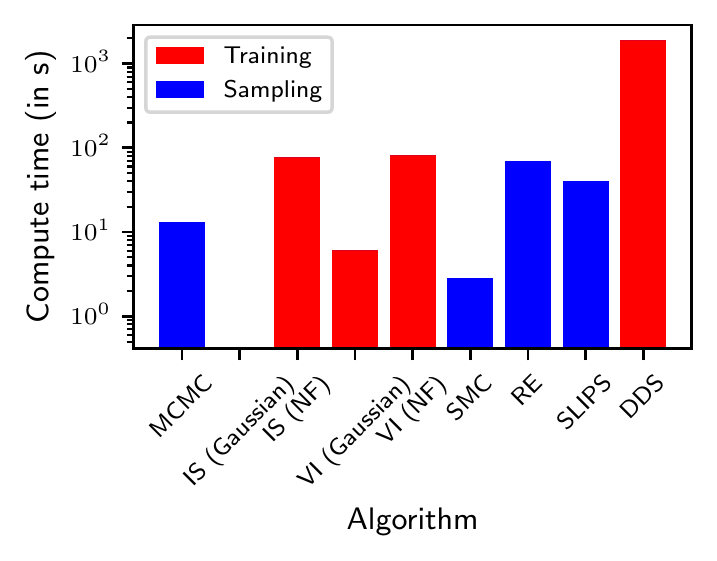}
    \caption{Average wall-clock computing time of each sampling algorithm. The results were averaged on $48 \times 25$ runs ($48$ Monte Carlo estimations for each of the $25$ different mode separation/dimension settings).}
    \label{fig:compute_time}
\end{figure}

\Cref{fig:compute_time} reports the average computing time of each algorithm. Note that, because of the on-the-fly discretization, the computational budget of SMC is stochastic. By way of indication, the budget of RE, SLIPS as well as the NF approaches would align with a 512 steps SMC algorithm.

